\documentclass{article}
\usepackage{ijcai23}

\usepackage{times}
\usepackage{soul}
\usepackage{xurl}
\usepackage[hidelinks]{hyperref}
\usepackage[utf8]{inputenc}
\usepackage[small]{caption}
\usepackage{graphicx}
\usepackage{amsmath}
\usepackage{amsthm}
\usepackage{booktabs}
\usepackage{algorithm}
\usepackage{algorithmic}
\usepackage[switch]{lineno}
\usepackage{amsmath}%
\usepackage{MnSymbol}%
\usepackage{wasysym}%
\usepackage{multirow}
\usepackage{hyperref}

\linenumbers

\title{A New Perspective for Shuttlecock Hitting Event Detection}

\author{
    Yu-Hsi Chen
    \affiliations
    Institute of Communications Engineering, National Tsing Hua University, Taiwan
    \emails
    yuhsi44165@gapp.nthu.edu.tw
}

\begin{document}

\nolinenumbers

\maketitle


\begin{abstract}
This article introduces a novel approach to shuttlecock hitting event detection. Instead of depending on generic methods, we capture the hitting action of players by reasoning over a sequence of images. To learn the features of hitting events in a video clip, we specifically utilize a deep learning model known as SwingNet. This model is designed to capture the relevant characteristics and patterns associated with the act of hitting in badminton. By training SwingNet on the provided video clips, we aim to enable the model to accurately recognize and identify the instances of hitting events based on their distinctive features. Furthermore, we apply the specific video processing technique to extract the prior features from the video, which significantly reduces the learning difficulty for the model. The proposed method not only provides an intuitive and user-friendly approach but also presents a fresh perspective on the task of detecting badminton hitting events. The source code will be available at \url{https://github.com/TW-yuhsi/A-New-Perspective-for-Shuttlecock-Hitting-Event-Detection}.
\end{abstract}

\section{Introduction}
In this task, we need to develop computer vision technology capable of automatically extracting comprehensive technical data from the broadcast video of badminton matches. This data includes crucial information about each shot, such as the precise timing, spatial positioning, player postures, and skill levels exhibited as the shuttlecock is struck during the match. It encompasses various elements, such as the shot's timing, ball's location, player responsible for the hit, swing posture, standing positions of both players, type of ball used, and the eventual winner. These extensive datasets serve as valuable resources for analyzing the technical and strategic aspects of badminton matches. The scoring system is based on individual rallies, where each rally is evaluated separately. Participants can earn a maximum of $1$ point per rally based on the accuracy of the rally's shot count and the precision of the predicted attributes. The final rankings are determined by the total scores, with the highest score securing the top position and the lowest score occupying the last spot.

\section{Related Work}
For this badminton competitions, it is common practice to use CoachAI~\cite{hsu2019coachai}, which is an AI-based coaching assistant system aimed at providing professional guidance and personalized advice in the field of sports. The system utilizes advanced machine learning and deep learning techniques to extract crucial information from athletes' data and videos, enabling comprehensive analysis and evaluation. Next, we will introduce the deep learning models and techniques utilized in CoachAI.

\subsection{TrackNetV2}
TrackNetV2~\cite{sun2020TrackNetV2} is an advanced computer vision model designed specifically for object tracking tasks. Building upon its predecessor, TrackNetV1, this updated version offers enhanced performance and improved accuracy in tracking objects across video sequences. It employs deep learning techniques, particularly convolutional neural networks (CNNs), to extract relevant features from consecutive frames of a video. These features are then used to predict the location and trajectory of a specified object in subsequent frames. By learning the spatiotemporal patterns and motion characteristics of the tracked object, TrackNetV2 can accurately follow its movements over time.

After obtaining the predicted badminton trajectories from TrackNetV2, further processing steps are performed, including removing extreme offset points, curve fitting, and interpolation. Once the data is prepared accordingly, roughly rally segmentation and event detection are applied to identify specific events within the processed videos.

\subsection{Court Detection}
Court detection in Coach AI plays a crucial role in enhancing the accuracy and effectiveness of the system, enabling it to provide valuable insights and actionable information based on the specific dynamics and requirements of the sports court.

\subsection{MoveNet}
MoveNet is a deep learning model designed for human pose estimation and tracking. It is specifically developed to accurately detect and track human body movements in real-time from video or camera input. This will be further used to analyze the BallType played by the players.

\subsection{OpenPose}
OpenPose is a computer vision library and framework for real-time multi-person keypoint detection and pose estimation. It enables the accurate estimation of human body poses, including the positions of body joints such as the head, shoulders, elbows, wrists, hips, knees, and ankles. Using OpenPose nodes in this competition allows for more precise identification of the location of the hitter and the defender.

\section{Methodology}
The methods we employ include video processing, SwingNet, ViT, YOLOv5, and TrackNetV2 deep learning models.

\subsection{Video Processing}
\begin{figure}
    \centering
    \includegraphics[width=8cm]{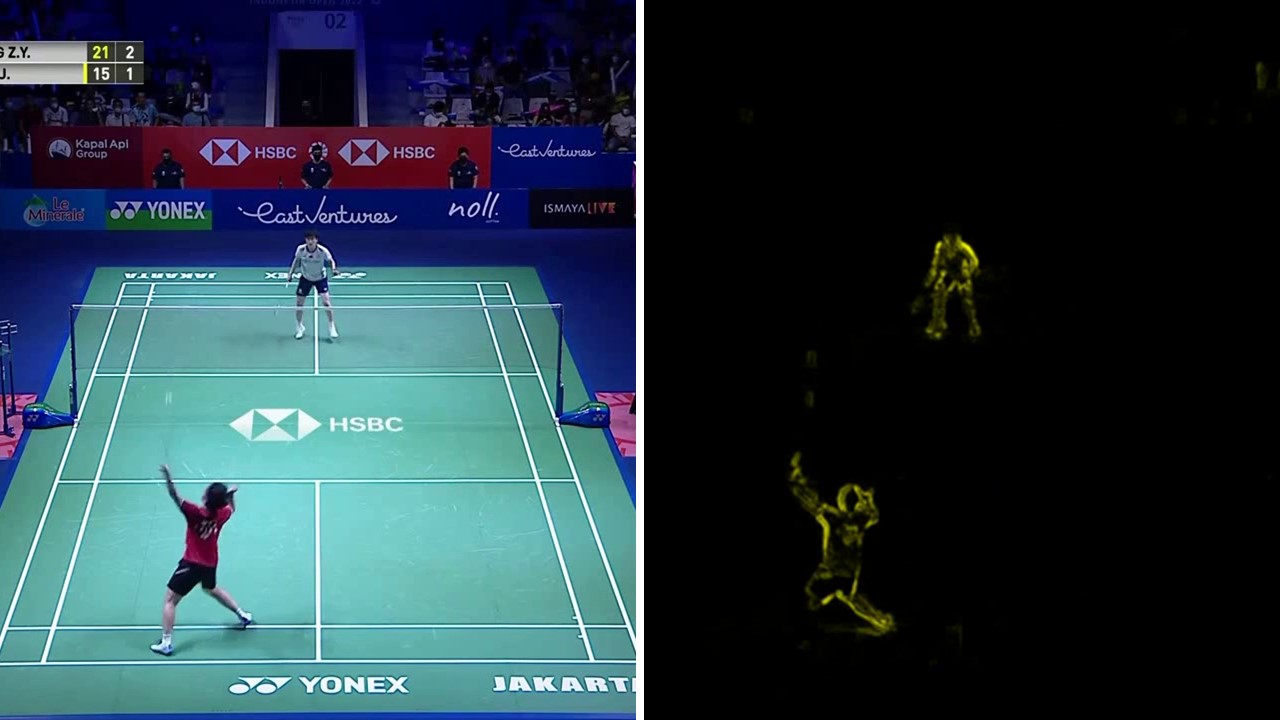}
    \caption[Comparison between the original and optical frames]{Comparison between the original and optical frames.}
    \label{fig:comp_origin_optical}
\end{figure}

Video processing is a critical step to improve the score in the entire task. In this process, we first calculate the optical flow images using the method of Optical Flow Calculation embedded in Reynolds Transport Theorem\footnote{\href{https://hdl.handle.net/11296/jf89b6}{A Prior Feature Enhanced Layer for Small Objects Detection and Tracking}}~\cite{bruhn2005lucas,harouna2017stochastic} to capture the prior features of the video, then, we remove the background information to achieve a mechanism similar to attention as shown in Figure~\ref{fig:comp_origin_optical}. And we will use the term "optical flow video without background" to specifically describe this type of video. Subsequently, we input these processed video into SwingNet for shuttlecock hitting event detection.

\subsection{SwingNet}
\begin{figure}
    \centering
    \includegraphics[width=8cm]{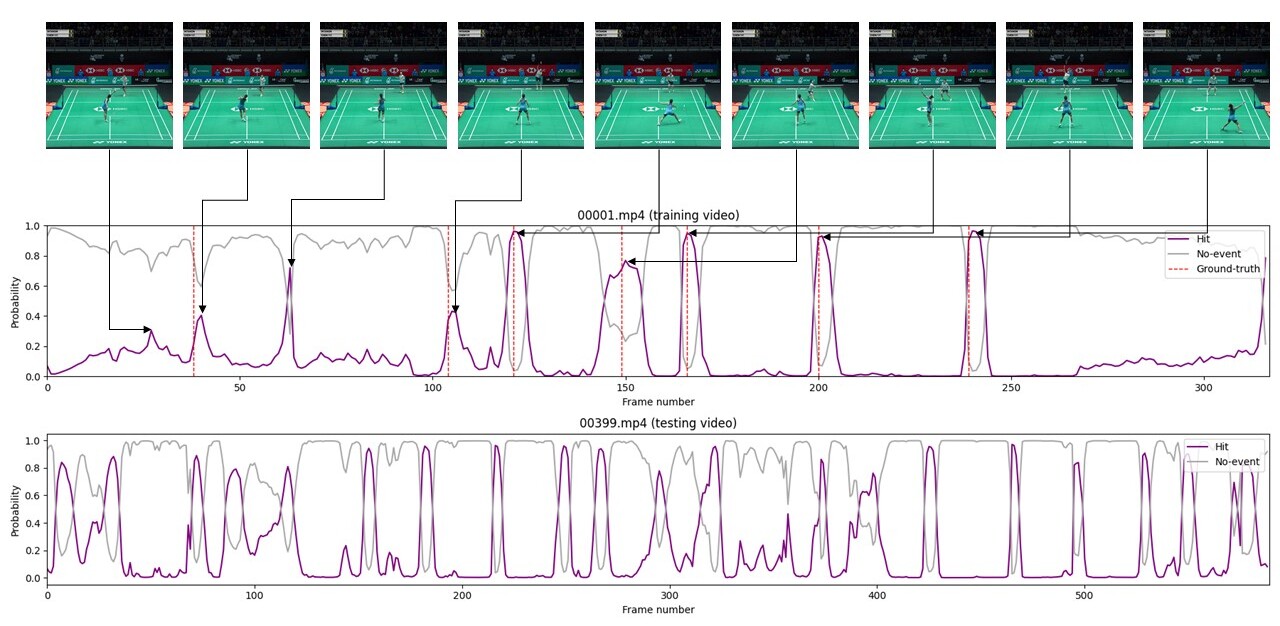}
    \caption[SwingNet performs event probability inference]{SwingNet performs event probability inference.}
    \label{fig:golfdb_event}
\end{figure}

SwingNet~\cite{mcnally2019golfdb} plays a crucial role in the field of computer vision and sports analytics, specifically in the domain of golf. Its application enables more efficient and detailed analysis of golf swings, contributing to the overall development and understanding of the sport. In here, we utilize the SwingNet, a deep learning model combining with the MobileNetV2 and bidirectional LSTM structure, for shuttlecock hitting event detection. That is, utilizing SwingNet enables us to extract the ShotSeq and HitFrame features for the desired csv file. Figure~\ref{fig:golfdb_event} demonstrates the inferred event probabilities for a video clip.

\subsection{Vision Transformer}
\begin{figure}[h]
    \centering
    \includegraphics[width=8cm]{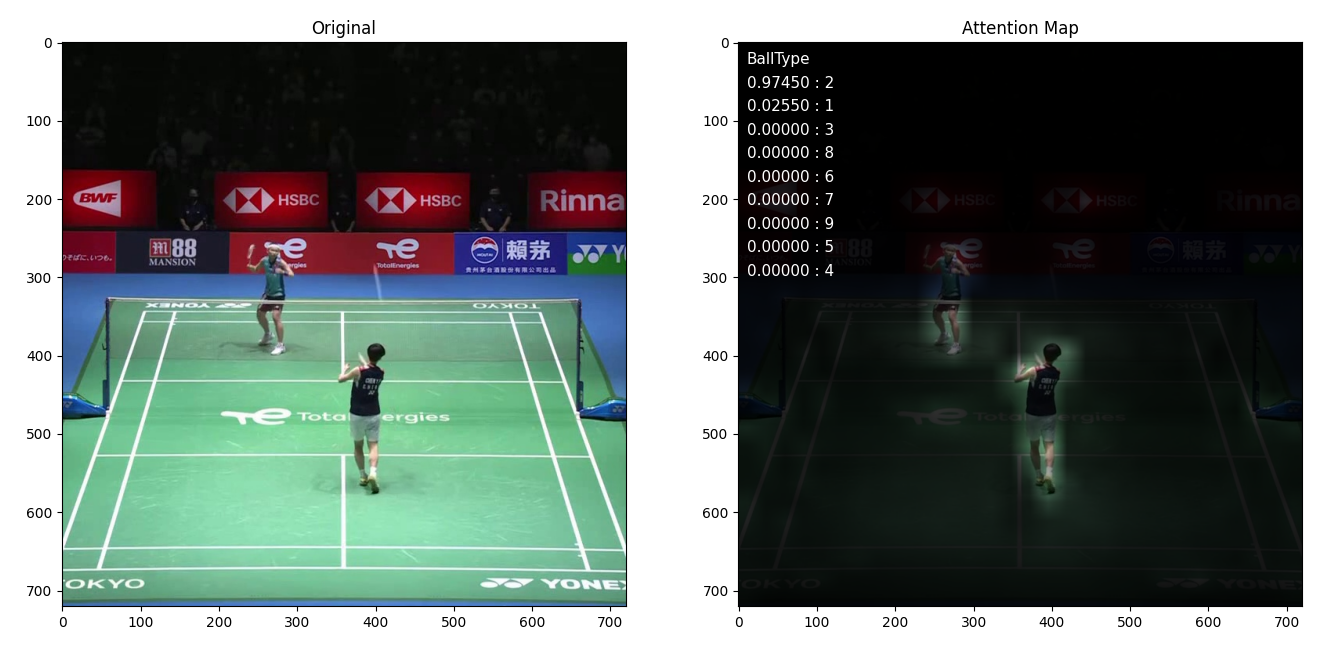}
    \caption[Attention map for BallType classification task]{Attention map for BallType classification task.}
    \label{fig:vit}
\end{figure}

\begin{figure}[h]
    \centering
    \includegraphics[width=8cm]{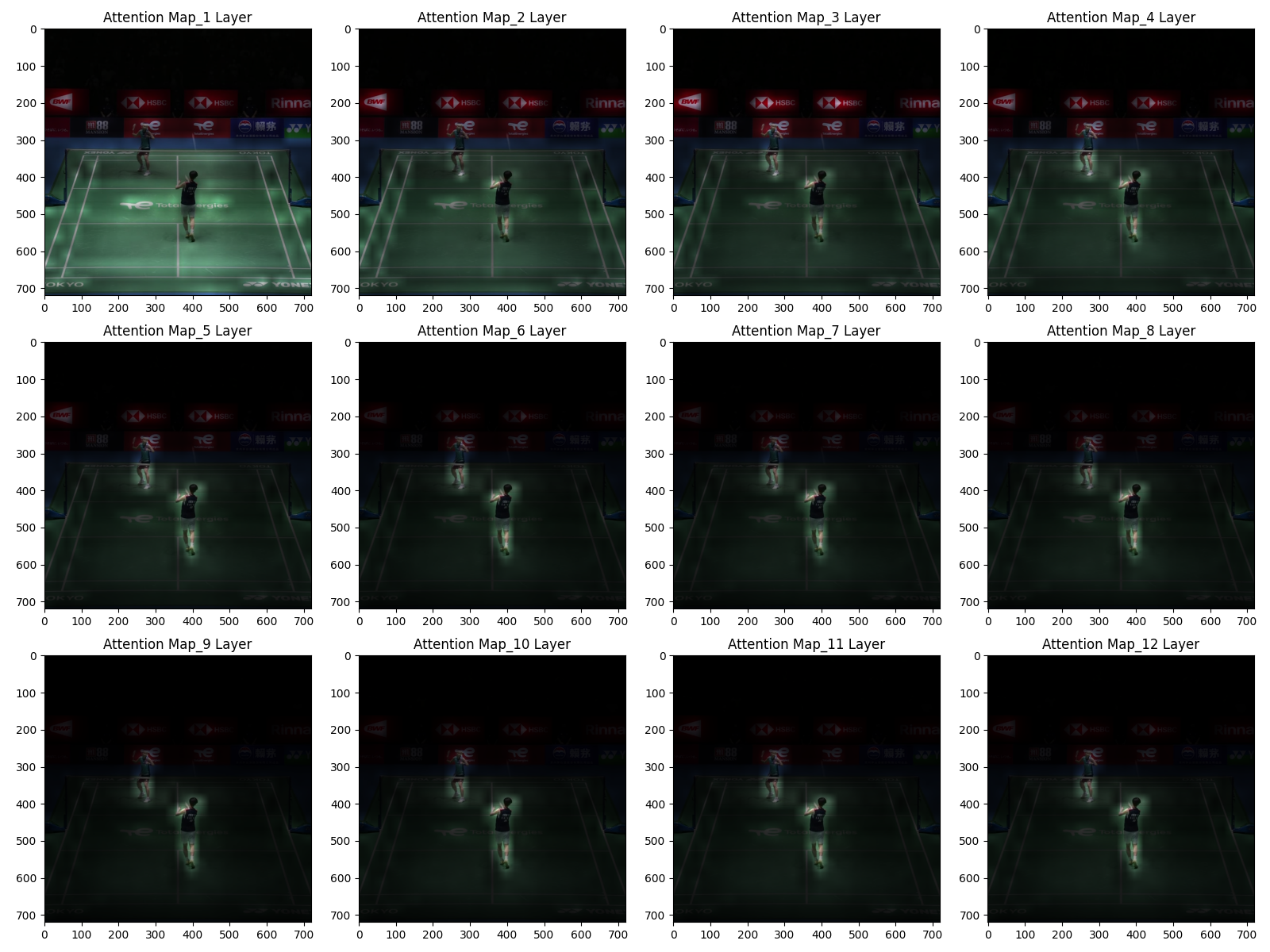}
    \caption[Visualization of attention maps for each layer]{Visualization of attention maps for each layer.}
    \label{fig:vit_layers}
\end{figure}

ViT, short for Vision Transformer~\cite{dosovitskiy2020image}, is a deep learning model that applies the transformer architecture to computer vision tasks. In contrast to the CNNs' approach, ViT adopts a distinct strategy by leveraging the transformer architecture, initially designed for tasks in natural language processing. Moreover, the strength of ViT lies in its attention mechanism, as depicted in Figure~\ref{fig:vit}, accompanied by the corresponding attention map illustrated in Figure~\ref{fig:vit_layers}. Here, we utilize ViT-B/16 to extract the information including Hitter, RoundHead, Backhand, BallHeight, BallType, and Winner from the videos. 

\subsection{YOLOv5}
The YOLO-series detector is a renowned family of object detection models. Specifically, YOLOv5~\cite{zhu2021tph} is a popular and highly efficient object detection model, which bilds upon the success of its predecessors, YOLOv1, YOLOv2, and YOLOv3, and introduces several improvements to achieve better performance in terms of accuracy and speed. We utilize YOLOv5m to extract the information for LandingX, LandingY, HitterLocationX, HitterLocationY, DefenderLocationX, and DefenderLocationY. 

\begin{figure}
    \centering
    \includegraphics[width=8cm]{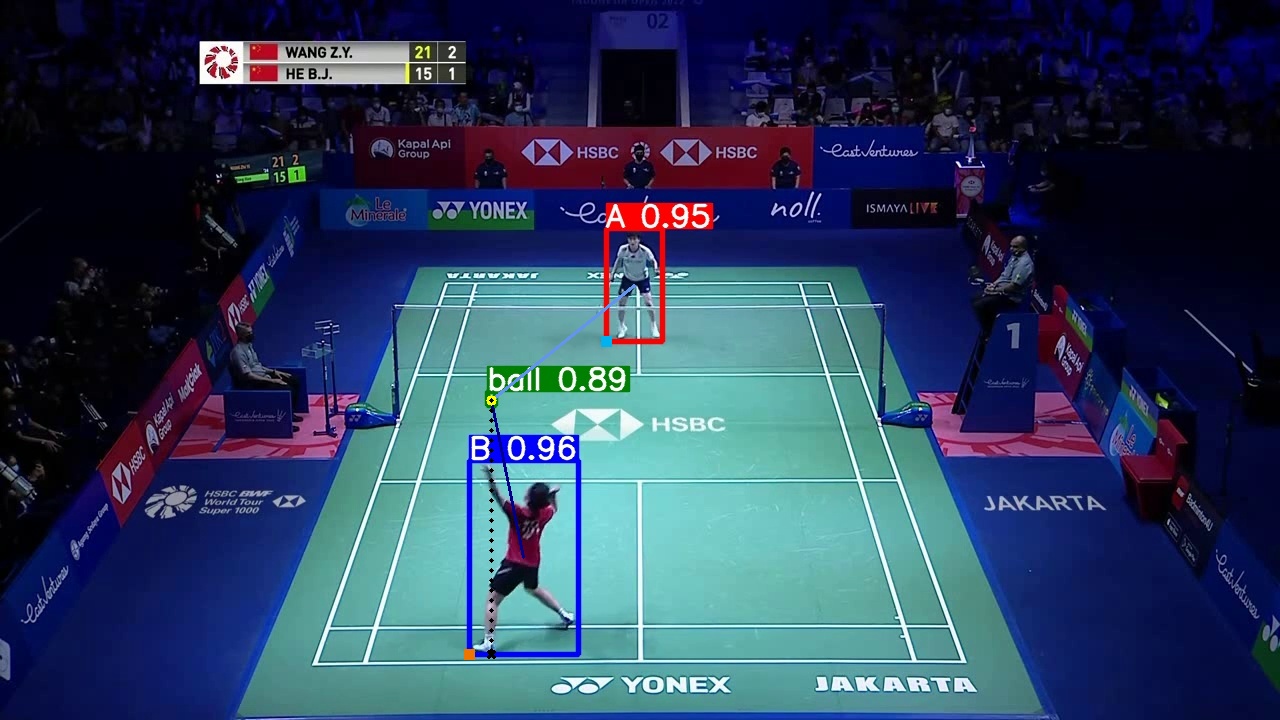}
    \caption[Detection result from YOLOv5m]{Detection result from YOLOv5m.}
    \label{fig:yolov5m}
\end{figure}

To begin with, we opted for YOLOv5m as our chosen detection model. The rationale behind not selecting newer models like YOLOv7 is based on my experience. While YOLOv7 may achieve higher scores on benchmark datasets, it lacks the desired level of robustness. Additionally, once we have obtained the detection results for both players and the ball, we integrate them with the Hitter prediction generated by ViT. We designate the output generated by ViT as the hitter, while the remaining player is assigned the role of the defender. As the landing result of the ball is the desired information, we utilize the y-value of the bottom-right corner of hitter's bounding box to represent the ball's landing y-value. The landing coordinate of the ball is indicated by the black cross symbol depicted in Figure~\ref{fig:yolov5m}. Next, we use the vertices of the detection boxes closest to the ball as the positions of the Hitter and the Defender for the AICUP competition, as illustrated by the light blue and orange rectangles in Figure~\ref{fig:yolov5m}.

\subsection{TrackNetV2}
The intention behind incorporating TrackNetV2 in this context is to compensate for the shortcomings of YOLO in detecting badminton. Nevertheless, there may still arise scenarios where both YOLOv5m and TrackNetV2 fail to detect the shuttlecock simultaneously. In that case, I will assign the coordinates of LandingX and LandingY as $(0,0)$.

\subsection{YOLOv8-pose}
\begin{figure}
    \centering
    \includegraphics[width=8cm]{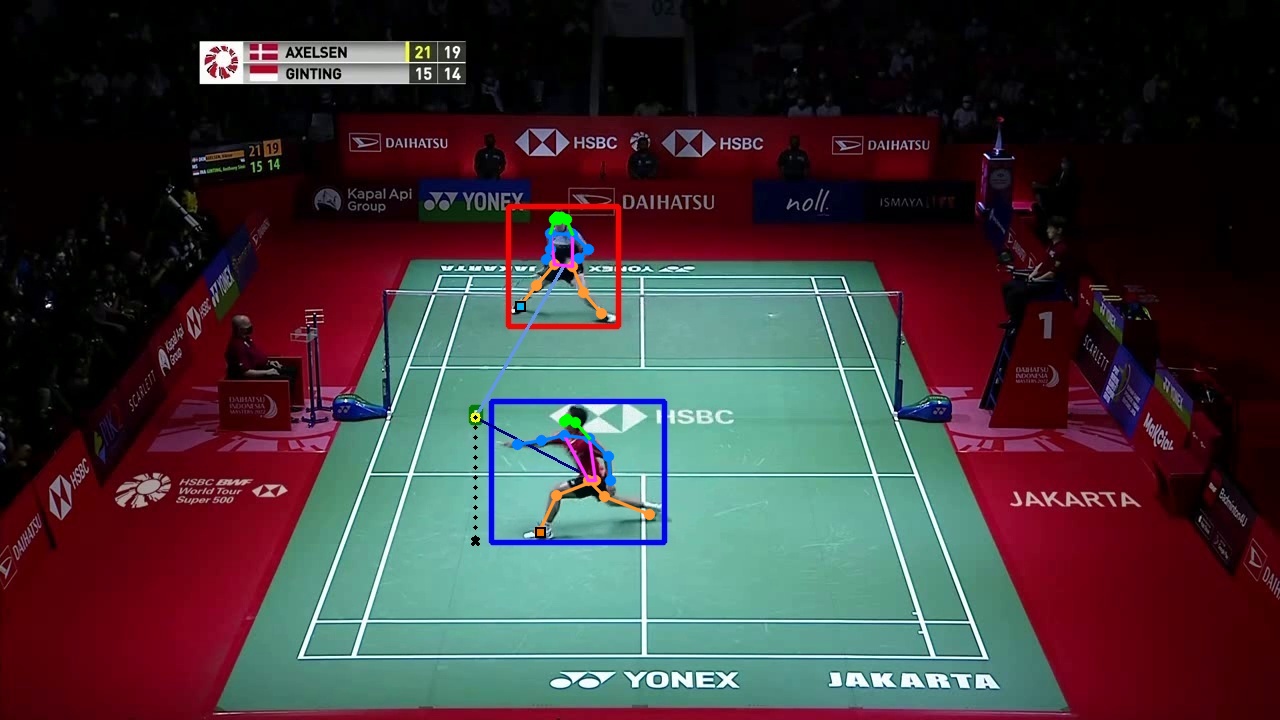}
    \caption[Detection results from YOLOv5m and YOLOv8-pose]{Detection results from YOLOv5m and YOLOv8-pose.}
    \label{fig:yolov8-pose}
\end{figure}

YOLOv8~\cite{Jocher_YOLO_by_Ultralytics_2023} is the latest version of YOLO by Ultralytics. Being at the forefront of advanced technology, YOLOv8 represents a SOTA model that builds upon the achievements of its predecessors. It incorporates innovative features and enhancements to deliver superior performance, versatility, and efficiency. Furthermore, YOLOv8 provides comprehensive support for a wide array of vision AI tasks, encompassing detection, segmentation, pose estimation, tracking, and classification. In the CodaLab competition, we adopt a different approach for determining the positions of the players. Instead of relying on the vertices of the detection boxes, we utilize the pose estimated by YOLOv8x-pose-p6 model to acquire the players' foot positions with enhanced accuracy. The image with the updated method is depicted in Figure~\ref{fig:yolov8-pose}.

\section{Results and Discussion}
This section encompasses the Evaluation Criteria, Experimental Results, and Discussion.

\subsection{Evaluation Criteria}
The scoring system is based on each individual video clip, with a maximum score of $1$ point per clip. The first step is to validate the number of shots within the rally video, with each row in the corresponding csv file representing one shot. If the predicted number of shots differs from the ground truth, the question will receive a score of $0$ points. However, if the number of shots is accurately predicted, a score of $0.1$ points will be awarded, and the evaluation will proceed to the next stage, which involves comparing the contents of the columns. In this stage, the maximum score for correctly matching column content is $0.9$ points.

When the number of shots is accurate for the test video, each shot is evaluated individually, and its score is determined based on the cumulative score of its columns. The average score of all the columns is then used as the shot's content score. The evaluation process begins by validating the HitFrame column. If the prediction error exceeds $2$ frames compared to the true value, the shot is awarded $0$ points in terms of content and the scoring proceeds to the next shot. However, if the prediction error is within or equal to $2$ frames, a score of $0.1$ points is given, and the remaining columns within the same shot are compared.
\begin{itemize}
\item Hitter: $0.1$ points if correct, $0$ points otherwise.
\item BallHeight: $0.1$ points if correct, $0$ points otherwise.
\item Landing: If the prediction error is not greater than $6$ pixels in the Euclidean distance, it is considered correct and is awarded $0.1$ points; otherwise, $0$ points.
\item HitterLocation: If the prediction error is not greater than $10$ pixels in the Euclidean distance, it is considered correct with $0.05$ points, otherwise $0$ points.
\item DefenderLocation: If the prediction error is not greater than $10$ pixels in the Euclidean distance, it is considered correct with $0.05$ points, otherwise $0$ points.
\item Backhand: $0.05$ points if correct, $0$ points otherwise.
\item RoundHead: $0.05$ points if correct, $0$ points otherwise.
\item BallType: $0.2$ points if correct, $0$ points otherwise.
\item Winner: $0.1$ points if correct, $0$ points otherwise. (Note: If it is not the last shot but is filled in, it will be considered incorrect.)
\end{itemize}
Assuming that there are R rally videos in the data set, and the $i$-th video has $S_i$ shots, the scoring formula is given below:
\begin{equation}
    \frac{1}{R}\sum\limits_{i=1}^{R} 1_{S_i = S_i^{pred}}(0.1 + ASS_i)
\end{equation}
in which $ASS_i$ (the Average Shot Score of content of the $i$-th rally video) is given by:
\begin{equation}
    ASS_i = \frac{1}{S_i}\sum\limits_{j=1}^{S_i}1_{|HitFrame_j - HitFrame_j^{pred}\leq 2}SS_j
\end{equation}
with $SS_j$ (the $j$-th Shot Score) given by:
\begin{equation}
    \begin{aligned}
        SS_j &= 0.1 + 0.1\times 1_{Hitter_j = Hitter_j^{pred}} \\
        &+ 0.1\times 1_{BallHeight_j = BallHeight_j^{pred}} \\
        &+ 0.1\times 1_{||Landing_j - Landing_j^{pred}<6} \\
        &+ 0.05\times 1_{||HitterLocation_j - HitterLocation_j^{pred}<10} \\
        &+ 0.05\times 1_{||DefenderLocation_j - DefenderLocation_j^{pred}<10} \\
        &+ 0.05\times 1_{Backhand_j = Backhand_j^{pred}} \\
        &+ 0.05\times 1_{RoundHead_j = RoundHead_j^{pred}} \\
        &+ 0.2\times 1_{BallType_j = BallType_j^{pred}} \\
        &+ 0.1\times 1_{Winner_j = Winner_j^{pred}}
    \end{aligned}
\end{equation}

\subsection{Experimental Results}
To begin with, we will discussing the hyperparameters and performance of SwingNet, as this aspect determines the upper limit of the overall score. This is due to the fact that any errors in the detection of ShotSeq and HitFrame would render subsequent feature accuracy irrelevant, as they would not contribute to the score. Subsequently, we will delve into the hyperparameters and performance analysis of the ViT-B/16 and YOLOv5m models. Consequently, the contribution of each model to the score will be illustrated at the conclusion of this section. Note that since the official pretrained TrackNetV2 and YOLOv8x-pose-p6 models are employed directly for inference, no further details will be provided here.

\subsubsection{Hyperparameters and Performance of SwingNet}
To enhance the performance of SwingNet in capturing key frames more effectively, we conducted numerous experiments focused on optimizing the hyperparameters. During the training phase, various parameters could be fine-tuned, including image size, sequence length, event ratio, the number of frozen layers, batch size, and iteration number. Additionally, during the inference phase, we can adjust parameters like image size, sequence length, probabilities for the quantiles, and the filter threshold for successive predicted events to further refine the results. After lots of trials, we have determined the hyperparameters that lead to improved predictions for the model. These optimized hyperparameters are summarized in Table~\ref{tab:hyperparameters_swingnet}.
\begin{table}
    \centering
    \begin{tabular}{lcc}
        \toprule
        Hyperparameters  & Train & Inference  \\
        \midrule
        image size & 180$\times$180 & 180$\times$180\\
        sequence length & 64 & 64\\
        number of frozen layers & 10 & 10\\
        batch size & 8 & 8\\
        event ratio & 0.0305 & --\\
        iteration number & 10000 & --\\
        probabilities for the quantiles & -- & 0.8\\
        filter threshold & -- & 3\\
        \bottomrule
    \end{tabular}
    \caption[Optimized hyperparameters for SwingNet]{Optimized hyperparameters for SwingNet.}
    \label{tab:hyperparameters_swingnet}
\end{table}

Once the hyperparameters of SwingNet have been determined, it becomes essential to examine how various video representations influence the detection results. We utilized three distinct types of input videos, including original video, optical flow video (Opt video), and optical flow video without background (Opt video w/o BG). Based on the information presented in Table~\ref{tab:SwingNet_different_videos}, it can be observed that utilizing the optical flow video without background yields the highest score compared to other types of videos.
\begin{table}
    \centering
    \begin{tabular}{cll}
        \toprule
        Video type  & Iter. $\backslash$ Loss & Score  \\
        \midrule
        \multirow{5}{*}{Original video}     & 2000 $\backslash$ 0.5821     & 0.0234 \\
             & 4000 $\backslash$    0.5074 & 0.0275  \\
             & 6000 $\backslash$    0.4622 & 0.0184  \\
             & 8000 $\backslash$    0.4269 & 0.0276  \\
             & 10000 $\backslash$   0.3979 & 0.0239  \\
        \midrule
        \multirow{5}{*}{Opt video}     & 2000 $\backslash$ 
             0.3799 & 0.0229 \\
             & 4000 $\backslash$    0.3325 & 0.0325  \\
             & 6000 $\backslash$    0.3037 & 0.0333  \\
             & 8000 $\backslash$    0.2839 & 0.0249  \\
             & 10000 $\backslash$   0.2664 & 0.0334  \\
        \midrule
        \multirow{5}{*}{Opt video w/o BG}     & 2000 $\backslash$ 0.3808     & 0.0274 \\
             & 4000 $\backslash$    0.3294 & 0.0331  \\
             & 6000 $\backslash$    0.2982 & \textbf{0.0349}  \\
             & 8000 $\backslash$    0.2760 & 0.0225  \\
             & 10000 $\backslash$   0.2582  & 0.0263  \\
        \bottomrule
    \end{tabular}
    \caption[Performance of SwingNet on different types of videos]{Performance of SwingNet on different types of videos.}
    \label{tab:SwingNet_different_videos}
\end{table}

Next, we made some architectural adjustments in an attempt to achieve higher scores. These adjustments included MobileNetV3 + bidirectional LSTM and MobileNetV2 + TCN. Before conducting the experiments, it was anticipated that both of these architectures would outperform the original one, as MobileNetV3~\cite{howard2019searching} and Temporal Convolution Network (TCN)~\cite{lea2017temporal} have been proven to be effective in other literature. However, compare to Table~\ref{tab:SwingNet_different_videos}, it can be clearly seen that the performance of the original architecture still remained the best. The performance of tuned architectures are shown in Table~\ref{tab:SwingNet_different_arch}.
\begin{table}
    \centering
    \begin{tabular}{cll}
        \toprule
        Architecture  & Iter. $\backslash$ Loss & Score  \\
        \midrule
        \multirow{2}{*}{MobileNetV3}     & 2000 $\backslash$ 0.4364     & 0.0136 \\
             & 4000 $\backslash$    0.3893 & 0.0293  \\
        +     & 6000 $\backslash$    0.3628 & 0.0162  \\
        \multirow{2}{*}{bidirectional LSTM}     & 8000 $\backslash$    0.3437 & 0.0253  \\
             & 10000 $\backslash$   0.3288 & 0.0115  \\
        \midrule
        \multirow{2}{*}{MobileNetV2}     & 2000 $\backslash$ 0.5417     & 0.0252 \\
             & 4000 $\backslash$    0.5065 & 0.0313  \\
        +     & 6000 $\backslash$    0.4816 & 0.0289  \\
        \multirow{2}{*}{TCN}     & 8000 $\backslash$    0.4613 & 0.0207  \\
             & 10000 $\backslash$   0.4440 & 0.0234  \\
        \bottomrule
    \end{tabular}
    \caption[Performance of tuned architectures]{Performance of tuned architectures.}
    \label{tab:SwingNet_different_arch}
\end{table}

Therefore, in this study, we primarily used optical flow without background as the main type of video for detection, and performed 5-fold inference. All the scores for the 5-fold inference are presented in Table~\ref{tab:SwingNet_G3}.
\begin{table}
    \centering
    \begin{tabular}{cccccc}
        \toprule
        Iter.$\backslash$fold & 1 & 2 & 3 & 4 & 5\\
        \midrule
        1000 & 0.0259& 0.0262& 0.0168& 0.027& 0.0263\\
        2000 & 0.0241& 0.0261& 0.0354& 0.034& 0.0339\\
        3000 & 0.0212& 0.0352& 0.037& 0.0318& \textbf{0.0426}\\
        4000 & 0.0275& 0.0266& 0.0389& 0.0372& 0.0399\\
        5000 & 0.0288& 0.027& 0.0289& 0.0366& 0.0351\\
        6000 & 0.0312& 0.0167& 0.0362& 0.0269& 0.0283\\
        7000 & 0.0233& 0.0372& 0.0144& 0.0326& 0.0319\\
        8000 & 0.0285& 0.0347& 0.0375& 0.0396& 0.0213\\
        9000 & 0.0236& 0.0237& 0.0387& 0.0279& 0.0226\\
        10000 & 0.0324& 0.0182& 0.0348& 0.0262& 0.0251\\
        \bottomrule
    \end{tabular}
    \caption[Scores obtained from 5-fold inference]{Scores obtained from 5-fold inference.}
    \label{tab:SwingNet_G3}
\end{table}

After multiple trials, the highest score obtained in the first two columns, as shown in Table~\ref{tab:SwingNet_G3}, is $0.0426$. In addition, we attempted to ensemble the detection results of SwingNet, but it did not effectively improve the score. Furthermore, it is notable that the maximum score for the first two columns can reach $0.2$ according to the evaluation criteria. This observation further emphasizes that SwingNet's predictions for the first two columns still have great potential for improvement.

Upon completing the first two columns, we will exhibit the hyperparameters employed for classification and detection tasks using ViT-B/16 and YOLOv5m, respectively.

\subsubsection{Hyperparameters of ViT-B/16 and YOLOv5m}
Based on numerous experiments conducted, we have determined the optimal hyperparameters that improve the performance of ViT-B/16 and YOLOv5m, as presented in Table~\ref{tab:hyperparameters_rest}. To enhance the inference performance of the ViT-B/16 and YOLOv5m models, the training strategy involves maximizing the image size. The main reason behind this setting is that larger image sizes have a tendency to uncover finer details, thereby significantly enhancing the value of both training and inference processes. In addition, I divided the training data into 5-fold for the classification tasks.
\begin{table}
    \centering
    \begin{tabular}{lcc}
        \toprule
        Hyperparameters & ViT-B/16 & YOLOv5m \\
        \midrule
        image size & 480 & 2880 \\
        optimizer & SGD & SGD \\
        learning rate & 3E-2 & 1E-2 \\
        loss function & CE & BCE \\
        batch size & 4 & 1\\
        iteration number & 10000 & --\\
        epochs & -- & 100 \\
        \bottomrule
    \end{tabular}
    \caption[Hyperparameters of ViT-B/16 and YOLOv5m]{Hyperparameters of ViT-B/16 and YOLOv5m.}
    \label{tab:hyperparameters_rest}
\end{table}

Once the hyperparameters of ViT-B/16 and YOLOv5m have been determined, it is important to delve into the stage of model inference. After completing the intact training phase, I will have a total of $30$ ViT-B/16 models for these classification tasks, specifically $5$ models for each. In the inference process, I have employed ensemble techniques, namely vote ensemble and mean ensemble. These approaches involve merging the predictions of multiple models by either averaging the probabilities or selecting the choice that receives the most votes, which aims to enhance the overall performance of the prediction. Table~\ref{tab:vit_ensemble} illustrates the comparison of scores achieved with and without the application of these ensemble techniques, using the previous predictions made by SwingNet as a baseline with a score of $0.0426$. 
\begin{table}
    \centering
    \begin{tabular}{lccc}
        \toprule
        Features$\backslash$Ensemble & fold1 & Vote & Mean \\
        \midrule
        Hitter & +0.0068 & +0.0068 & +0.0068 \\
        RoundHead & +0.0033 & +0.0033 & +0.0033 \\
        Backhand & +0.0031 & +0.0031 & +0.0031 \\
        BallHeight & +0.0066 & +0.0066 & \textbf{+0.0067} \\
        BallType & +0.0089 & \textbf{+0.0091} & +0.0089 \\
        Winner & +0.0005 & +0.0005 & \textbf{+0.0006} \\
        \bottomrule
    \end{tabular}
    \caption[Ablation study of ensemble techniques]{Ablation study of ensemble techniques.}
    \label{tab:vit_ensemble}
\end{table}

It can be observed that there is only a slight difference between using and not using ensemble techniques. Furthermore, the variations among different ensemble techniques are also small from Table~\ref{tab:vit_ensemble}. The main factor contributing to this observation is the notable disparity in probabilities predicted by ViT-B/16. As illustrated in Figure~\ref{fig:vit}, the model assigns a remarkably high probability of $0.9745$ to the second category, while the probability of the ball belonging to the next highest category only $0.0255$, the disparity between these probabilities is substantial. Consequently, the utilization of ensemble techniques does not significantly impact the classification results, as the dominant probability already indicates a clear classification preference. Therefore, we can train only one ViT-B/16 model for each feature during classification tasks, rather than five models. This approach greatly reduces the time and resources required for training and inference.

Now, let's analyze the individual contributions of different models to their respective features. Firstly, we will use SwingNet to obtain the predicted answers for ShotSeq and HitFrame. Subsequently, we utilize ViT-B/16 for predicting the features related to the classification task. Furthermore, we employ the YOLOv5m model with TrackNetV2 to detect the position of the players and the trajectory of the badminton. Moreover, we utilized the YOLOv8x-pose-p6 model to estimate the poses of players, resulting in a more accurate detection of their positions as shown in Figure~\ref{fig:yolov8-pose}. These steps allow us to generate the complete submission file. The corresponding contribution table can be found in Table~\ref{tab:Contribution_table}.
\begin{table}
    \centering
    \begin{tabular}{lll}
        \toprule
        Features & Models & Score \\
        \midrule
        ShotSeq & \multirow{2}{*}{SwingNet} & \multirow{2}{*}{0.0426} \\
        \&HitFrame & & \\
        Hitter & ViT & 0.0494 (+0.0068) \\
        RoundHead & ViT & 0.0527 (+0.0033) \\
        Backhand & ViT & 0.0558 (+0.0031) \\
        BallHeight & ViT & 0.0625 (+0.0067) \\
        LandingX & YOLOv5m & \multirow{2}{*}{0.0625 (+0.0000)} \\
        \&LandingY & +TrackNetV2 & \\
        HitterLocationX & \multirow{4}{*}{\shortstack[l]{YOLOv5m\\ +YOLOv8x\\\quad-pose-p6}} & \multirow{4}{*}{0.0630 (+0.0005)} \\
        \&HitterLocationY & & \\
        \&DefenderLocationY & & \\
        \&DefenderLocationY & & \\
        BallType & ViT & 0.0721 (+0.0091) \\
        Winner & ViT & 0.0727 (+0.0006) \\
        \bottomrule
    \end{tabular}
    \caption[The distinct contributions of various models to the features]{The distinct contributions of various models to the features.}
    \label{tab:Contribution_table}
\end{table}

\subsection{Discussion}
In this section, we will primarily address four key points of discussion. The first one is the extensive GPU memory resources required when using the method proposed in this article for shuttlecock hitting event detection with SwingNet. The main reason is that the model reads in $64$ frames of video at once for evaluation. For example, if we input $64$ full-color images with resolution of $720p$, the model needs to accommodate a size of $64\times24\times1280\times720/8/1000\approx 169$GB. Due to the limited $6$GB memory of my GPU, the maximum image resolution we can use for training and testing is only $180\times 180$. I believe that if there is sufficient GPU memory available, utilizing the method proposed in this article and increasing the image size could be a promising approach. Compared to analyzing the trajectory of the shuttlecock, this method is more intuitive and can overcome situations where the shuttlecock is obstructed by the player.

The second point is that, according to the rules of badminton, we can infer that the information in the Hitter column will alternate between $A$ and $B$ $(A\rightarrow B\rightarrow A\rightarrow B\cdots)$. This implies that there will never be more than two consecutive instances of either $A$ or $B$. Consequently, by predicting the first hitter, we can determine the order of each hitter in a video. It is crucial to emphasize that when employing the alternative showup method, the accuracy of the predicted first hitter becomes paramount. Any inaccuracies in this prediction may result in a complete reversal of the order for every hitter in the entire video.

The third point to consider is that by exclusively focusing on the hitter during the classification process, it is possible to enhance the performance of the classification problem. That is, instead of utilizing the entire image for classification as depicted in Figure~\ref{fig:vit}, one can solely utilize the area occupied by the hitter. Implementing this method has the potential to decrease the defender's propensity to confuse hitter's information, such as RoundHead, Backhand, and BallType.

Lastly, it is worth noting that in this competition, several features were not utilized while training, such as LandingX, LandingY, HitterLocationX, HitterLocationY, DefenderLocationX, DefenderLocationY. Without fully utilizing the available information, I believe there will be a certain degree of misalignment. Henceforth, we will analyze the unused information with the aim of leveraging it to aid in both the training and inference stages.

\section{Summary}
This article proposes a new perspective for the shuttlecock hitting event detection task, which involves using a deep learning model to learn the meaning conveyed by the image sequence, and thus capturing the desired information from the video. Furthermore, the video processing method presented in this article can effectively reduce the difficulty of model learning, resulting in better performance in keyframe detection task. In the future, we will continue to explore adjustments to the model architecture in order to construct a more precise method for obtaining keyframes. This will be combined with suitable image processing techniques to further enhance the overall framework.

\section*{Acknowledgments}
I am delighted to have the opportunity to integrate the models used in this competition and engage in deeper discussions about their structures and mechanisms. I am also grateful to my family and friends for their unwavering support and encouragement throughout this period, which enabled me to fully immerse myself in the contests and generate innovative ideas. I truly appreciate it.

\bibliographystyle{named}
\bibliography{ijcai23}

\end{document}